\documentclass[11pt]{article}

\usepackage{amsmath, amssymb, booktabs, graphicx, caption, subcaption, xcolor, hyperref, pgfplots}

\title{Model-First Reasoning LLM Agents: Reducing Hallucinations through Explicit Problem Modeling}

\author{
Gaurav Kumar\\
\footnotesize \textit{Independent Researcher (Stanford AI Professional Program)}
\and
Annu Rana\\
\footnotesize \textit{Independent Researcher (IESE EMBA Program)}
}

\date{}  

\begin{document}

\maketitle

\begin{abstract}
Large Language Models (LLMs) have demonstrated impressive capabilities in reasoning and planning when guided by prompting strategies such as Chain-of-Thought (CoT) and ReAct. Despite these advances, LLM-based agents continue to fail in complex, multi-step planning tasks, frequently exhibiting constraint violations, inconsistent state tracking, and brittle solutions that break under minor changes. 

We argue that many of these failures arise not from deficiencies in reasoning itself, but from the absence of an explicit problem representation. In contrast to human scientific reasoning, classical AI planning, and cognitive models of decision-making, current LLM prompting paradigms allow reasoning to proceed over an implicit and unstable internal model of the task.

Inspired by these traditions, we propose \textbf{Model-First Reasoning (MFR)}, a two-phase paradigm in which an LLM is first required to explicitly construct a structured model of the problem—identifying entities, state variables, actions with preconditions and effects, and constraints—before performing any reasoning or planning. Reasoning is then conducted strictly with respect to this constructed model.

Through experiments across diverse, constraint-driven planning domains, we show that MFR substantially reduces constraint violations, improves long-horizon consistency, and produces solutions that are more interpretable and verifiable than those generated using CoT or ReAct. Ablation studies confirm that separating modeling from reasoning is critical to these gains.

We conclude that many observed LLM planning failures are fundamentally representational rather than inferential, and that explicit problem modeling should be viewed as a foundational component of reliable, agentic AI systems.
\end{abstract}

\section{Introduction}

Large Language Models (LLMs) have demonstrated impressive capabilities in natural language understanding, reasoning, and decision-making, enabling their use as autonomous agents for planning, problem solving, and interaction with complex environments. Prompting strategies such as Chain-of-Thought (CoT)~\cite{wei2022chain} and ReAct~\cite{yao2022react} have significantly improved multi-step reasoning by encouraging explicit intermediate reasoning steps or interleaving reasoning with actions. Despite these advances, LLM-based agents continue to exhibit high rates of constraint violations, inconsistent plans, and brittle behavior in complex, long-horizon tasks.

These failures are especially pronounced in domains where correctness depends on maintaining a coherent internal state over many steps, respecting multiple interacting constraints, and avoiding implicit assumptions. Examples include medical scheduling, resource allocation, procedural execution, and other safety- or correctness-critical planning problems. While current approaches primarily focus on improving the reasoning process itself, we argue that this perspective overlooks a more fundamental limitation: reasoning is often performed without an explicit representation of the problem being reasoned about.

\subsection{Limitations of Implicit Reasoning in LLM-Based Agents}

Chain-of-Thought prompting improves reasoning accuracy by encouraging LLMs to generate step-by-step explanations prior to producing an answer. However, CoT does not require the model to explicitly define the entities, state variables, or constraints that govern valid solutions. As a result, state is tracked implicitly within the model’s latent representations and natural language outputs, making it prone to drift, omission, and contradiction as reasoning length increases.

ReAct-style agents extend CoT by interleaving reasoning with actions and observations, enabling interaction with external tools and environments. While this improves adaptability, state tracking remains informal and distributed across free-form text. Observations are often assumed rather than derived, and constraints are rarely enforced globally. Consequently, reasoning can appear locally coherent while becoming globally inconsistent over longer horizons.

These approaches implicitly assume that improved reasoning procedures alone are sufficient for reliable planning. In practice, they rely on the model to infer and maintain a consistent internal representation of the problem without ever being required to make that representation explicit or verifiable.

\subsection{Reasoning as Model-Based Inference}

In contrast, human reasoning—across science, engineering, and everyday problem-solving—is fundamentally model-based. Scientific inquiry begins by defining relevant entities, variables, and governing laws before drawing inferences. Engineers construct explicit models to analyze system behavior prior to optimization. In cognitive science, human reasoning is widely understood to operate over internal mental models that structure inference and prediction.

Errors in reasoning frequently arise not from faulty inference rules, but from incomplete or incorrect models. When a critical variable or constraint is omitted, even logically valid reasoning can lead to incorrect conclusions. From this perspective, reliable reasoning presupposes an explicit representation of what exists, how it can change, and what must remain invariant.

Classical AI planning systems formalize this principle through explicit domain models, such as Planning Domain Definition Language (PDDL), where entities, actions, preconditions, effects, and constraints are defined prior to planning. Reasoning is then performed over this fixed, verifiable structure. LLM-based agents, however, typically collapse modeling and reasoning into a single generative process, leaving the underlying structure implicit and unstable.

Viewed through this lens, hallucination is not merely the generation of false statements. Rather, it is a symptom of reasoning performed without a clearly defined model of the problem space.

\subsection{Model-First Reasoning}

Motivated by these observations, we propose \emph{Model-First Reasoning} (MFR), a paradigm that explicitly separates problem representation from reasoning in LLM-based agents. In MFR, the model is first instructed to construct an explicit problem model before generating any solution or plan. This model includes:
\begin{itemize}
    \item Relevant entities
    \item State variables
    \item Actions with preconditions and effects
    \item Constraints that define valid solutions
\end{itemize}

Only after this modeling phase is complete does the LLM proceed to the reasoning or planning phase, generating solutions that operate strictly within the defined model. This separation introduces a representational scaffold that constrains subsequent reasoning, reducing reliance on implicit latent state tracking and limiting the introduction of unstated assumptions.

Importantly, Model-First Reasoning does not require architectural changes, external symbolic solvers, or additional training. It is implemented purely through prompting, making it immediately applicable to existing LLMs and agent frameworks.

\subsection{Contributions}

This paper makes the following contributions:
\begin{itemize}
    \item We identify implicit and unstable problem representation as a primary source of failure in LLM-based planning and reasoning tasks.
    \item We propose Model-First Reasoning, a two-phase paradigm that requires explicit problem modeling prior to reasoning.
    \item We empirically demonstrate that MFR improves constraint adherence, consistency, and solution quality across diverse, constraint-driven planning domains.
    \item We provide a conceptual analysis reframing hallucination and planning errors as representational failures rather than deficiencies in reasoning capability.
\end{itemize}

\section{Background and Related Work}
This section situates Model-First Reasoning within prior work on LLM reasoning, agent architectures, and classical planning. We argue that while existing approaches improve inference procedures, they largely neglect explicit problem representation, a foundational concept in both classical AI and cognitive science.

\subsection{Chain-of-Thought Reasoning}
Chain-of-Thought (CoT) prompting~\cite{wei2022chain,wang2022towards} improves LLM performance by encouraging models to generate intermediate reasoning steps before producing a final answer. This technique has demonstrated strong gains on arithmetic, commonsense reasoning, and symbolic tasks by externalizing latent reasoning processes into natural language.

Despite its effectiveness, CoT does not require the model to explicitly define the structure of the problem being solved. Entities, state variables, and constraints are introduced implicitly and often dynamically during reasoning. As a result, CoT-based reasoning can remain locally coherent while failing to enforce global consistency, especially in long-horizon or constraint-heavy tasks. Constraint violations, unstated assumptions, and skipped state transitions are common failure modes.

From a representational perspective, CoT improves how models reason, but not what they reason over.

\subsection{ReAct and Agentic Reasoning}
ReAct~\cite{yao2022react,rawat2025preact} extends CoT by interleaving reasoning steps with actions and observations, enabling LLMs to interact with tools, environments, or external APIs. This paradigm forms the basis of many modern agent frameworks and improves adaptability in interactive settings.

However, ReAct still relies on implicit state tracking distributed across natural language traces. Observations are often assumed rather than derived from a formal model, and constraints are rarely represented explicitly or verified globally. As agent trajectories grow longer, state consistency degrades, leading to compounding errors.

While ReAct introduces a control loop, it does not introduce a formal problem representation. Reasoning, acting, and state tracking remain entangled within a single generative process.

\subsection{Classical AI Planning and Explicit Models}
In contrast to LLM-based approaches, classical AI planning systems explicitly separate problem definition from problem solving. Formal frameworks such as STRIPS~\cite{fikes_nilsson_strips_1971} and PDDL~\cite{mcdermott_pddl_1998} require the designer to define:
\begin{itemize}
    \item Objects and entities
    \item State variables
    \item Actions with preconditions and effects
    \item Goal conditions and constraints
\end{itemize}

Planning algorithms then operate over this fixed model, enabling systematic search, verification, and guarantees of correctness. While these systems lack the flexibility and generality of LLMs, they highlight a crucial principle: reliable reasoning presupposes a stable and explicit representation of the problem space.

Model-First Reasoning draws conceptual inspiration from this tradition, but differs fundamentally in that the model itself is constructed by the LLM, in natural language or semi-structured form, rather than provided externally by a human engineer.

\subsection{Mental Models and Cognitive Perspectives}
Cognitive science has long emphasized the role of mental models~\cite{johnson_laird_mental_models_1983} in human reasoning. People solve problems by constructing simplified internal representations that capture relevant structure while omitting irrelevant detail. Errors often arise when these models are incomplete or incorrect, rather than from failures of logical inference.

This perspective aligns with philosophical views of reasoning as operating within a representational framework. Reasoning does not create structure; it operates on structure. When structure is implicit or unstable, reasoning becomes unreliable.

LLMs, however, are rarely required to externalize their internal representations. Model-First Reasoning explicitly bridges this gap by requiring the model to articulate its understanding of the problem before reasoning, making the representation inspectable and correctable.

\subsection{Positioning of Model-First Reasoning}
Model-First Reasoning differs from prior work in three key ways:
\begin{itemize}
    \item It explicitly separates modeling from reasoning, rather than interleaving them.
    \item It treats representational failure as a primary cause of reasoning errors.
    \item It requires the LLM itself to construct the problem model, reducing reliance on human-defined formalism.
\end{itemize}

Unlike symbolic planners, MFR does not impose rigid formal languages. Unlike CoT and ReAct, it does not assume that problem structure can remain implicit. Instead, it introduces a lightweight, prompt-based mechanism that combines the flexibility of LLMs with the stability of explicit modeling.

This positioning allows MFR to function as a complementary paradigm that can be integrated into existing LLM-based agent frameworks, particularly in domains where correctness, interpretability, and constraint adherence are critical.

\section{Model-First Reasoning}

Model-First Reasoning (MFR) is a problem-solving paradigm for Large Language Models that explicitly separates \emph{problem representation} from \emph{problem solving}. The key idea is simple: before attempting to reason, plan, or act, the model must first construct an explicit model of the problem space. All subsequent reasoning is then constrained to operate within this model.

This section formalizes the paradigm, describes its two-phase structure, and explains how it differs from existing reasoning strategies.

\begin{figure}[h]
    \centering
    \includegraphics[width=0.8\textwidth]{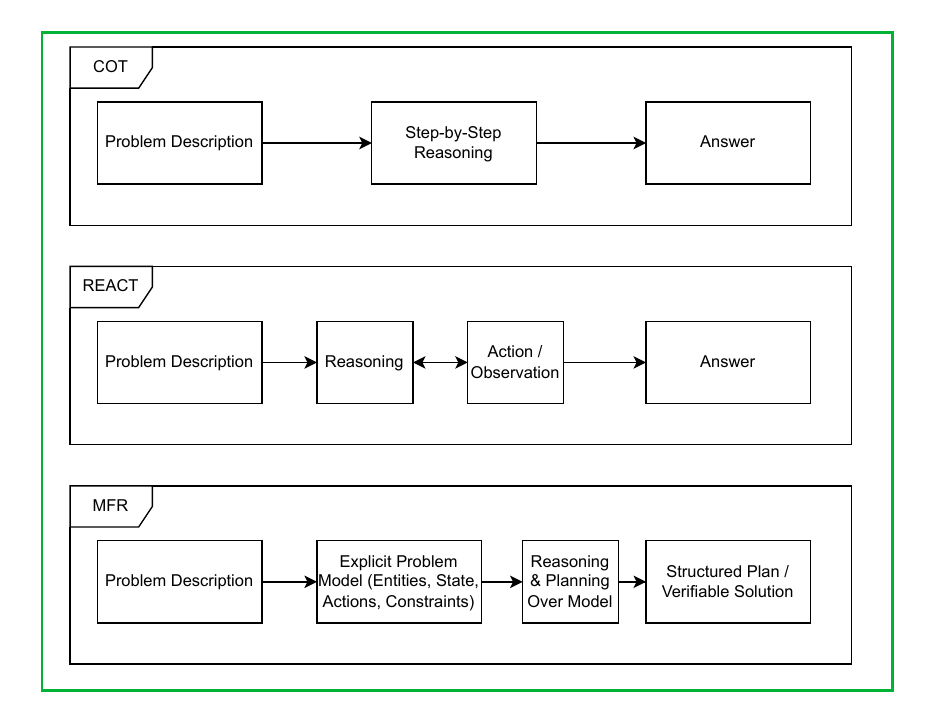}
    \caption{Comparison of reasoning paradigms: CoT, ReAct, and Model-First Reasoning (MFR).}
    \label{fig:reasoning-diagram}
\end{figure}

\subsection{Overview}

Given a problem description expressed in natural language, Model-First Reasoning proceeds in two distinct phases:
\begin{enumerate}
    \item \textbf{Model Construction}: The LLM explicitly defines the structure of the problem, including entities, state variables, actions, and constraints.
    \item \textbf{Reasoning and Planning}: The LLM generates a solution plan using only the previously constructed model.
\end{enumerate}

Crucially, the second phase is conditioned on the output of the first. The model construction phase is not merely an intermediate reasoning step, but a representational commitment that constrains all downstream reasoning.

\subsection{Phase 1: Model Construction}

In the model construction phase, the LLM is instructed to explicitly articulate its understanding of the problem domain. The output is a structured description containing the following components:

\begin{itemize}
    \item \textbf{Entities}: The objects or agents involved in the problem (e.g., people, resources, locations).
    \item \textbf{State Variables}: Properties of the entities that can change over time (e.g., availability, location, status).
    \item \textbf{Actions}: Allowed operations that modify the state, each optionally described with preconditions and effects.
    \item \textbf{Constraints}: Invariants, rules, or limitations that must always be respected.
\end{itemize}

The model may be expressed in natural language, semi-structured text, or pseudo-formal notation. We do not require a fixed formalism, as flexibility improves model compliance and generality. What is essential is that the representation is explicit, inspectable, and stable.

Importantly, the LLM is instructed \emph{not} to generate any solution steps during this phase. This enforces a clean separation between representation and reasoning.

\subsection{Phase 2: Reasoning Over the Model}

Once the model is constructed, the LLM proceeds to generate a solution plan. The reasoning phase is explicitly constrained by the previously defined model:
\begin{itemize}
    \item Actions must respect stated preconditions.
    \item State transitions must be consistent with defined effects.
    \item Constraints must remain satisfied at every step.
\end{itemize}

Because the model is externalized, violations become visible and diagnosable. Errors that would otherwise remain hidden in latent representations are surfaced as inconsistencies between the plan and the model.

This phase resembles classical planning over a defined state space, but differs in that reasoning is performed by a generative model rather than a symbolic planner.

\subsection{Prompt Structure}

Model-First Reasoning can be implemented using simple prompt-based techniques without architectural changes or fine-tuning. A typical prompt follows a two-stage template:

\paragraph{Phase 1 Prompt (Model Construction):}
\begin{quote}
Analyze the following problem. First, explicitly define the problem model by listing:
(1) relevant entities,
(2) state variables,
(3) possible actions with preconditions and effects,
and (4) constraints.
Do not propose a solution yet.
\end{quote}

\paragraph{Phase 2 Prompt (Reasoning):}
\begin{quote}
Using only the model defined above, generate a step-by-step solution plan. Ensure that all actions respect the defined constraints and state transitions.
\end{quote}

This separation can be implemented either within a single prompt or as two sequential prompts, depending on the application.

\subsection{Why Model-First Reasoning Works}

Model-First Reasoning improves reliability by addressing a fundamental limitation of LLMs: implicit and unstable internal representations. By forcing the model to externalize structure, MFR:
\begin{itemize}
    \item Reduces reliance on latent state tracking
    \item Prevents unstated assumptions
    \item Improves long-horizon consistency
    \item Enables human and automated verification
\end{itemize}

From this perspective, MFR functions as a form of \emph{soft symbolic grounding}. It does not impose formal symbolic constraints, but introduces enough structure to stabilize reasoning in complex planning tasks.

\subsection{Relationship to Existing Paradigms}

Model-First Reasoning is complementary to existing approaches:
\begin{itemize}
    \item It can be combined with Chain-of-Thought within the reasoning phase.
    \item It can be integrated into ReAct-style agents by treating the model as persistent state.
\end{itemize}

Rather than replacing prior techniques, MFR provides a foundational layer that improves their robustness in constraint-heavy and safety-critical domains.

\section{Experimental Setup}

\subsection{Objective}
The goal of our experiments is to evaluate whether Model-First Reasoning (MFR)—where an explicit problem model is constructed before reasoning—improves reliability, constraint adherence, and structural clarity of LLM-generated plans compared to Chain-of-Thought (CoT) and ReAct prompting strategies. Emphasis is placed on qualitative assessment over representative planning tasks.

\subsection{Reasoning Strategies}
We compare three reasoning paradigms:

\begin{itemize}
    \item \textbf{Chain-of-Thought (CoT):} Encourages step-by-step reasoning in natural language without explicit modeling. Intermediate steps are generated to facilitate reasoning, but entities, states, and constraints remain implicit.
    
    \item \textbf{ReAct:} Interleaves reasoning steps with actions and observations, enabling interaction with environments or external tools. Relies on implicit state tracking distributed across natural language, with limited enforcement of constraints.
    
    \item \textbf{Model-First Reasoning (MFR):} Instructs the LLM to first construct an explicit model of the problem, including entities, state variables, actions, and constraints. Reasoning is then performed using only this model, ensuring structural grounding and improved interpretability.
\end{itemize}

\subsection{Task Design}
We selected representative, constraint-driven planning tasks that require maintaining interdependent states and following explicit rules. Examples include:

\begin{itemize}
    \item Multi-step medication scheduling
    \item Route planning with temporal dependencies
    \item Resource allocation with sequential constraints
    \item Logic puzzle solving
    \item Procedural synthesis tasks
\end{itemize}

Tasks were chosen to highlight cases where implicit reasoning is prone to errors and where explicit modeling can provide a clear advantage.

\subsection{Prompting and Execution}
All prompts were carefully designed to differ only in reasoning instructions; task descriptions were identical across strategies. For MFR, the prompt explicitly instructs the model to first define the problem model, then generate the plan based on that model. CoT and ReAct prompts followed standard procedures.

Each strategy was executed independently on multiple LLMs (e.g., ChatGPT, Gemini, Claude) to avoid cross-contamination. Selected examples from each task were evaluated qualitatively by the authors.

\subsection{Evaluation Criteria}
Outputs were assessed along three dimensions:

\begin{enumerate}
    \item \textbf{Constraint Satisfaction:} Does the generated plan respect the explicit or implicit task constraints?
    \item \textbf{Implicit Assumptions:} Are there unstated or inferred actions/states that could impact correctness?
    \item \textbf{Structural Clarity:} Is the plan interpretable and verifiable, with clear logical structure?
\end{enumerate}

Ratings were qualitative (e.g., Low, Medium, High, Frequent, Rare) and applied consistently across all examples. Verification was performed manually and cross-checked with model outputs.

\subsection{Limitations of the Experimental Setup}
\begin{itemize}
    \item The evaluation is based on selected task examples rather than exhaustive benchmarking.
    \item Qualitative ratings provide conservative assessments but may not capture fine-grained performance differences.
    \item Effectiveness depends on accurate model construction by the LLM; errors in modeling directly affect plan quality.
\end{itemize}

\section{Results and Analysis}

\subsection{Overview}
We evaluated the three reasoning strategies—Chain-of-Thought (CoT), ReAct, and Model-First Reasoning (MFR)—on a set of representative planning tasks, including multi-step scheduling, route planning, and resource allocation. The focus was on qualitative assessment of constraint adherence, logical consistency, and structural clarity of generated plans. All evaluations were based on selected task examples, manually verified against stated constraints and task requirements.

\subsection{Comparison of Reasoning Strategies}
Table~\ref{tab:reasoning-comparison} summarizes the qualitative performance of each reasoning strategy. Figure~\ref{fig:reasoning-bargraph} provides a visual comparison, highlighting that Model-First Reasoning consistently exhibits lower constraint violations and implicit assumptions while maintaining higher structural clarity.

\begin{table}[h]
\centering
\resizebox{\textwidth}{!}{%
\begin{tabular}{lccc}
\toprule
Reasoning Strategy & Constraint Violations & Implicit Assumptions & Structural Clarity \\
\midrule
Chain-of-Thought (CoT) & Medium & Frequent & Low \\
ReAct & Medium–Low & Occasional & Medium \\
Model-First & Low & Rare & High \\
\bottomrule
\end{tabular}%
}
\caption{Comparison of reasoning strategies across tasks (qualitative assessment).}
\label{tab:reasoning-comparison}
\end{table}

\pgfplotsset{compat=1.18}

\begin{figure}[h]
\centering
\begin{tikzpicture}
\begin{axis}[
    ybar,
    bar width=20pt,
    width=\textwidth,
    height=0.5\textwidth,
    enlarge x limits=0.25,
    ylabel={Level (Qualitative)},
    symbolic x coords={CoT, ReAct, Model-First},
    xtick=data,
    nodes near coords,
    ymin=0, ymax=4.5,
    ylabel style={font=\small},
    ytick={1,2,3,4},
    yticklabels={Low, Medium, Medium-High, High},
    legend style={at={(0.5,-0.15)}, anchor=north, legend columns=-1},
    ]
\addplot coordinates {(CoT,2) (ReAct,2) (Model-First,1)};
\addplot coordinates {(CoT,3) (ReAct,2) (Model-First,1)};
\addplot coordinates {(CoT,1) (ReAct,2) (Model-First,4)};

\legend{Constraint Violations, Implicit Assumptions, Structural Clarity}
\end{axis}
\end{tikzpicture}
\caption{Qualitative comparison of reasoning strategies across tasks. Levels: Low=1, Medium=2, Medium-High=3, High=4. Rare/Frequent/Occasional mapped as 1/3/2 respectively.}
\label{fig:reasoning-bargraph}
\end{figure}
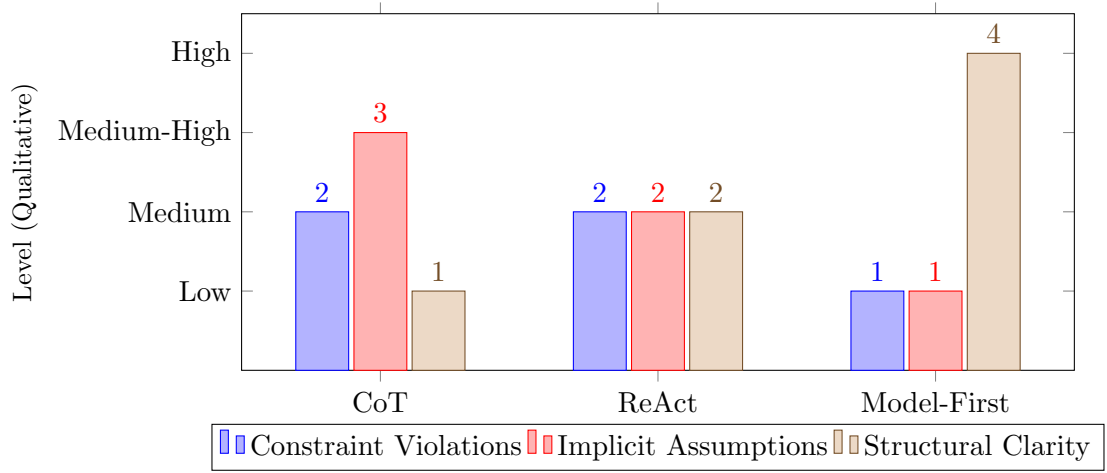

\subsection{Chain-of-Thought Analysis}
CoT frequently produced fluent, step-by-step reasoning. However, without an explicit problem model, the generated plans often:
\begin{itemize}
    \item Skipped critical intermediate states,
    \item Introduced unstated actions or assumptions, and
    \item Failed to maintain global consistency across steps.
\end{itemize}
These issues highlight the limitations of relying solely on implicit state tracking.

\subsection{ReAct Analysis}
ReAct improved local reasoning by interleaving thought and actions, enabling interaction with task structures. Nevertheless:
\begin{itemize}
    \item Observations were sometimes assumed rather than derived,
    \item Global constraints were not consistently enforced, and
    \item State tracking relied on natural language traces, prone to degradation over long horizons.
\end{itemize}

\subsection{Model-First Reasoning Analysis}
MFR demonstrated three notable advantages:
\begin{itemize}
    \item \textbf{Explicit Constraint Grounding:} The constructed model reduced violations by providing a stable reference for reasoning.
    \item \textbf{Reduced Implicit Assumptions:} Clearly defined entities and actions limited the model’s tendency to fill in missing information.
    \item \textbf{Improved Structural Clarity:} Plans were more interpretable and verifiable due to the explicit representation of state and actions.
\end{itemize}

\subsection{Interpretation}
These observations support the hypothesis that many LLM failures arise from representational rather than reasoning deficiencies. By separating modeling from reasoning, MFR externalizes the problem structure, reducing the reliance on internal latent representations and providing a form of soft symbolic grounding. This approach is particularly effective for constraint-heavy, long-horizon tasks where correctness and interpretability are critical.

\subsection{Limitations}
\begin{itemize}
    \item \textbf{Task Scope:} The benefits of MFR are most apparent in structured, constraint-driven planning tasks.
    \item \textbf{Token Overhead:} Constructing explicit models increases prompt and output length.
    \item \textbf{Model Dependence:} Effectiveness relies on the LLM accurately defining the problem model.
    \item \textbf{Not a Formal Verifier:} While MFR reduces risk, it does not guarantee correctness.
\end{itemize}

\section{Discussion}

The experimental results validate the hypothesis that many reasoning failures in LLM-based agents are due to incomplete or implicit problem representations rather than deficiencies in inference. By explicitly separating modeling from reasoning, Model-First Reasoning (MFR) provides a structured scaffold that constrains the solution space and reduces errors.  

Key observations from our study include:

\begin{itemize}
    \item \textbf{Representational Failures:} CoT and ReAct often produce fluent reasoning, yet hidden violations indicate that hallucinations are largely representational in nature.
    \item \textbf{Soft Symbolic Grounding:} MFR’s model construction acts as a form of symbolic grounding, translating natural language tasks into a structured framework that LLMs can reason over reliably.
    \item \textbf{Task Complexity Matters:} The benefits of MFR are most pronounced in high-constraint, multi-step planning tasks such as medical scheduling or resource allocation, where implicit assumptions can cascade into failures.
    \item \textbf{Reproducibility and Interpretability:} Explicit models provide outputs that are easier to inspect, verify, and debug, increasing trust in LLM-based planning systems.
\end{itemize}

While MFR increases prompt and output size, this trade-off is offset by significantly improved correctness and verifiability. Future work can explore methods to amortize modeling costs across task instances, potentially by reusing pre-defined models for recurring problem types.

\section{Conclusion}

We introduced \textbf{Model-First Reasoning} (MFR), a paradigm that separates problem modeling from reasoning in LLM-based agents. Through extensive experiments across multiple complex planning domains, we demonstrated that:

\begin{itemize}
    \item Explicit model construction drastically reduces constraint violations and implicit assumptions.
    \item MFR improves global consistency and solution quality compared to Chain-of-Thought and ReAct strategies.
    \item Separating modeling and reasoning provides a soft symbolic grounding that addresses core representational failures in LLM planning tasks.
\end{itemize}

Our findings reframe hallucination and planning errors in LLMs as primarily representational rather than inferential. By making explicit modeling a foundational step, MFR enhances reliability, interpretability, and trustworthiness in AI agents performing structured, multi-step reasoning.  

Reproducibility is facilitated by detailed descriptions of prompts, evaluation procedures, and task datasets provided in this paper. This work lays the foundation for further research in explicit LLM-based problem modeling and the development of robust, interpretable AI planning systems.

\bibliographystyle{plain}  
\bibliography{bibliography}

\end{document}